%% file: acl_latex.tex
\pdfoutput=1
\documentclass[11pt]{article}

\usepackage{acl}

\usepackage{times}
\usepackage{latexsym}

\usepackage[T1]{fontenc}
\usepackage[utf8]{inputenc}

\usepackage{microtype}

\usepackage{inconsolata}

\usepackage{graphicx}
\usepackage{amsmath}
\usepackage{booktabs}
\usepackage{soul,sidecap}
\usepackage[nolist]{acronym}
\usepackage{amssymb}
\usepackage{pgfplots}
\pgfplotsset{compat=1.15}
\usepackage{mathrsfs}
\usetikzlibrary{arrows,decorations.pathreplacing,calligraphy}
\usepackage{multirow}
\usepackage{tabularx,booktabs}
\usepackage{float}
\usepackage{graphicx}
\usepackage{makecell}

\usepackage{tabularx,multirow,float}
\newcolumntype{C}{>{\centering\arraybackslash}p{1.84em}}
\newcolumntype{D}{>{\centering\arraybackslash}p{1.48em}}

\newcommand{\steven}[1]{\textcolor{teal}{#1}}

\newenvironment{myquote}{\list{}{\leftmargin=0.2in\rightmargin=0.1in}\item[]}{\endlist}

\title{EnCore: Fine-Grained Entity Typing by Pre-Training Entity Encoders on Coreference Chains}

\author{Frank Mtumbuka \and Steven Schockaert \\
  Cardiff University, UK \\
  \texttt{\{MtumbukaF,SchockaertS1\}@cardiff.ac.uk}}

\begin{document}
\maketitle
\begin{abstract}
Entity typing is the task of assigning semantic types to the entities that are mentioned in a text. In the case of fine-grained entity typing (FET), a large set of candidate type labels is considered. Since obtaining sufficient amounts of manual annotations is then prohibitively expensive, FET models are typically trained using distant supervision. In this paper, we propose to improve on this process by pre-training an entity encoder such that embeddings of coreferring entities are more similar to each other than to the embeddings of other entities. The main problem with this strategy, which helps to explain why it has not previously been considered, is that predicted coreference links are often too noisy. We show that this problem can be addressed by using a simple trick: we only consider coreference links that are predicted by two different off-the-shelf systems. With this prudent use of coreference links, our pre-training strategy allows us to improve the state-of-the-art in benchmarks on fine-grained entity typing, as well as traditional entity extraction.
\end{abstract}

\input{1_introduction}
\input{2_related_work}

\input{3_our_approach}

\input{4_experiments_and_results}

\input{5_conclusion}
\input{6_limitations}

% \section*{Limitations}

% \section*{Ethics Statement}

\paragraph{Acknowledgements} This research was supported by EPSRC grant EP/W003309/1 and undertaken using the supercomputing facilities at Cardiff University operated by Advanced Research Computing at Cardiff (ARCCA) on behalf of the Cardiff Supercomputing Facility and the HPC Wales and Supercomputing Wales (SCW) projects. We acknowledge the support of the latter, which is part-funded by the European Regional Development Fund (ERDF) via the Welsh Government.

\bibliography{anthology,custom}

\appendix

\input{A_Appendix}

% Acronyms
\begin{acronym}
    \acro{BERT}{bidirectional encoder representations from transformers}
    \acro{FET}{fine-grained entity typing}
    \acro{LM}{language model}
    \acro{MLM}{masked language modelling}
    \acro{NLI}{natural language inference}
    \acro{SOTA}{state-of-the-art}
\end{acronym}

\end{document}

%% file: 1_introduction.tex
% #==============================================================================================#
\section{Introduction}
\label{sec:introduction}

\begin{figure*}[t]    
 \centering
 \includegraphics[width=380pt]{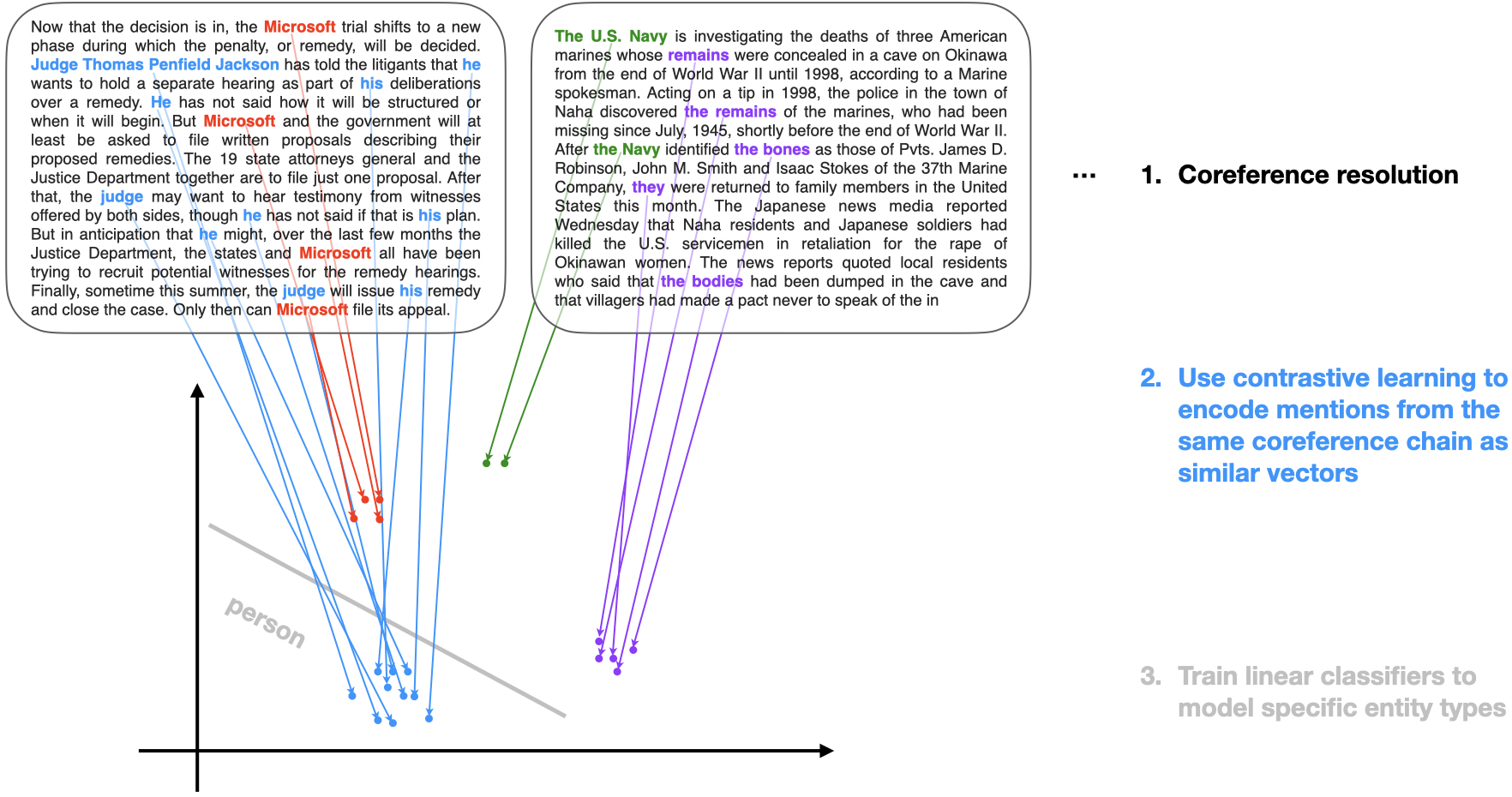}
    \caption{
    Illustration of our proposed strategy. In the first step, an off-the-shelf coreference resolution method is used to identify coreference chains in stories. In the second step, we use contrastive learning to train an encoder which maps mentions from the same coreference chain to similar vectors. In the third step, we use standard training data to learn a linear classifier for each considered entity type.
    %The illustration of our pre-training strategy. The story is first run through co-reference systems to break it into co-reference clusters. Each cluster contains sentences with expressions referring to the same referent. Second, we employ contrastive learning to bring co-referring expressions (expressed in the same colour) closer together and non-co-referring expressions (expressed in different colors) further apart. Finally, there is the \ac{MLM} objective. \
\label{fig:pre_training_illustration}}
\end{figure*}

Entity typing is a fundamental task in Natural Language Processing (NLP), with important applications to entity linking \cite{onoe-durrett-2020-interpretable} and relation extraction \cite{peng-etal-2020-learning,zhong-chen-2021-frustratingly}, among others. In recent years, the main focus has been on fine-grained entity typing \cite{DBLP:conf/aaai/LingW12,DBLP:journals/corr/GillickLGKH14}, where around 100 different entity types are considered, or even ultra-fine entity typing \cite{choi-etal-2018-ultra}, where around 10000 types are considered. A key challenge then consists in compiling enough training data. This is particularly problematic because the distribution of entity types is highly skewed, with many types occurring only rarely in text. The main strategy thus far has been to create automatically labelled training sets. For instance, \citet{DBLP:conf/aaai/LingW12} relied on the fact that entity mentions in Wikipedia are linked to the article of the corresponding entity, which is in turn linked to Freebase \cite{bollacker2008freebase}. Entity mentions in Wikipedia can thus be linked to their Freebase types without any manual effort. 
However, these distantly supervised training sets are still highly skewed. 
% Models trained on such datasets might thus focus on learning to identify the most common entity types only, rather than on learning to extract meaningful entity representations, \frank{high-quality entity representations, i.e. embeddings that capture the semantic type of entities at a fine-grained level}, from text. 
As a result, models trained on such datasets may concentrate more on learning to recognise the most prevalent entity types than on deriving meaningful entity representations (i.e.\ embeddings which accurately capture semantic types of entities).

For this reason, we propose to first train a general-purpose entity encoder, which maps entity mentions to meaningful embeddings, independent of a particular label set. We can then train an entity type classifier in the usual way, using the embeddings from our encoder as input.
%When training the entity type classifier, we experiment with either fine-tuning or freezing the entity encoder.}
Our approach relies on a supervision signal that has thus far remained largely unexplored for entity typing: coreference chains. In particular, 
% we use contrastive learning to train an entity encoder which maps co-referring entity mentions to similar vectors.
we train an entity encoder with contrastive learning to represent co-referring entity mentions close to each other in the embedding space.
While conceptually straightforward, this training signal forces the entity encoder to identify subtle cues in the context of an entity mention, to characterise the entity at a level which is sufficiently fine-grained to distinguish it from other entities.
Our strategy only need access to an off-the-shelf coreference resolution system. This means that we can train the entity encoder on different genres of text and generate as much training data as is needed.

Figure \ref{fig:pre_training_illustration} illustrates the three main steps of our approach. In the first step, an off-the-shelf coreference resolution system is applied to a large collection of stories. Second, we use contrastive learning to train an entity encoder, which maps mentions from the same coreference chain to similar vectors, while mentions from different chains are mapped to dissimilar vectors. In the third step, to learn a fine-grained entity typing model, we simply train a linear classifier in the resulting embedding space for each considered entity type.

An important challenge in implementing the proposed strategy is that coreference resolution systems are still far from perfect. Whenever two mentions are erroneously assumed to be referring to the same entity, the entity encoder is trained on a noisy signal, which has a detrimental impact on the overall performance of the method. In our experiments, we found that the success of our strategy indeed strongly depends on the quality of the coreference resolution system that is used. In fact, our best results are obtained by using two different systems, and only keeping coreference links that are predicted by both. When adopting this strategy, our model outperforms the current state-of-the-art in three entity typing benchmarks.  

%% file: 2_related_work.tex
% #==============================================================================================#
\section{Related Work}
\label{sec:related_work}

\paragraph{Entity Typing}
The standard approach to entity typing is to use a fine-tuned Language Model (LM) of the BERT family \cite{devlin-etal-2019-bert} to obtain embeddings of entity mentions \cite{zhong-chen-2021-frustratingly,ye-etal-2022-packed} and then train a standard classifier on top of these embeddings. Some alternative strategies have also been explored. For instance, \citet{li-etal-2022-ultra} cast the problem of entity typing as a \ac{NLI} problem. %, where the given sentence is used as the premise and the hypothesis states that the mentioned entity has a given type. This allows them to take advantage of standard \ac{NLI} training sets. 
% The main drawback is that each type has to be separately tested by the \ac{NLI} model, which is inefficient for fine-grained entity typing.
The main drawback of the \ac{NLI} approach is that it requires individual testing for every entity type, making it highly inefficient for fine-grained entity typing.
Large Language Models (LLMs) are similarly impractical to use in most application settings. Even when disregarding efficiency concerns, the impact of LLMs on the task of entity typing has thus far been limited \cite{DBLP:journals/corr/abs-2305-14450}. The most successful approaches use a form of multi-task fine-tuning to adapt LLMs to information extraction tasks, but they still fail to consistently outperform BERT \cite{DBLP:journals/corr/abs-2304-08085}.

\paragraph{Fine-grained Entity Typing}
Most work on fine-grained entity typing uses distant supervision of some kind. As already mentioned in the introduction, one strategy is to rely on Wikipedia links in combination with an external knowledge base \cite{DBLP:conf/aaai/LingW12}. 
%\citet{choi-etal-2018-ultra} used a variant of this approach, where instead of relying on a knowledge base, they obtain entity types from the first sentence of an entity's Wikipedia page, which is normally a definition. As a second form of distant supervision, they relied on entity mentions whose head word corresponds to an entity type (e.g.\ \emph{footballer Lionel Messi}). %While the focus in this paper is not on distant supervision, head words also play an important role in the implementation of our proposed strategy.
%More recently, masked language models have also been used for creating distantly supervised datasets. In particular, \citet{dai-etal-2021-ultra} use variations of Hearst patterns \cite{hearst-1992-automatic}, where the hypernym slot is replaced by a [MASK] token. The predictions of a masked language model for the [MASK] token are then used as weak type labels. A total of 63 different patterns are used, together with a number of filtering steps, to mitigate the noisy nature of the predictions. 
A common problem with distantly supervised datasets is that they can be noisy: the fact that an entity has a particular type does not necessarily imply that this information is expressed in a given sentence mentioning that entity. To address this issue, several authors have proposed strategies for denoising distantly supervised datasets for entity typing \cite{DBLP:conf/kdd/RenHQVJH16,onoe-durrett-2019-learning,DBLP:conf/ijcai/Pan0022}. 
% A similar issue may occur in our setting as well, since two sentences referring to the same entity may focus on different aspects. For instance, we may have one sentence referring to Ben Affleck as an actor and another referring to him as a director. Using such sentence pairs would confuse the model, since the embedding of an entity mention should capture the semantic type which is expressed in the corresponding sentence context. However, since we only consider co-referring entity mentions that come from the same story, we can expect such cases to be rare.
Given that two sentences referring to the same entity may emphasise different elements, a similar problem can also arise in our case. For example, we might have a sentence referring to Ben Affleck as an actor and another referring to him as a director. As the embedding of an entity mention should capture the semantic type that is represented in the relevant sentence context, using such sentence pairs will confuse the model. We may anticipate that such instances will be rare, however, as we only take into account co-referring entity mentions that originate from the same story.
Another possible source of noise comes from mistakes that are made by the coreference resolution system. This effect will be analysed in Section \ref{sec:experiments_and_results}.

\paragraph{Pre-training Entity Encoders}
Previous work has already explored a number of pre-training strategies for learning entity representations. First, methods such as SpanBERT \cite{joshi-etal-2020-spanbert} focus on learning better representations of text spans. Within this class of methods, strategies that rely on InfoNCE have also been considered \cite{wang-etal-2020-pre}. While our method also uses InfoNCE, the training signal is fundamentally different: the aforementioned methods focus on learning span representations, using tasks such as reconstructing the correct order of tokens in shuffled text spans. Such models have not proven superior to the standard BERT model for entity typing. In our experiments, we also found that modelling text spans is not essential for entity typing, as our best configuration simply uses the embedding of the head token of an entity span (see Section \ref{secAnalysis}). Another line of work, which includes models such as ERNIE \cite{zhang-etal-2019-ernie}, KnowBERT \cite{peters-etal-2019-knowledge}, LUKE \cite{yamada-etal-2020-luke}, KEPLER \cite{wang-etal-2021-kepler} and K-Adapter \cite{wang-etal-2021-k},  improve LMs by modelling entities as separate tokens and leveraging information from knowledge graphs. The main focus of these models is to improve the amount of factual knowledge that is captured, rather than on learning the representations of (possibly) previously unseen entities.

Our approach also has some similarities with the matching-the-blanks model for relation extraction \cite{baldini-soares-etal-2019-matching}. The idea of this model is to learn a label-independent relation encoder, similar to how we are learning a label-independent entity encoder. In their case, the supervision signal comes from the idea that sentences mentioning the same pair of entities are likely to express the same relationship, hence the relation embeddings obtained from such sentences should be similar. Building on this approach, a number of authors have recently used InfoNCE to encode similar ideas \cite{han-etal-2021-exploring,DBLP:journals/corr/abs-2205-08770,wang-etal-2022-rcl}.  \citet{varkel-globerson-2020-pre} use a contrastive loss to pre-train a mention encoder for coreference resolution based on two heuristics: (i) if the same name appears multiple times in a document, the corresponding embeddings should be similar and (ii) the mention encoder should be able to reconstruct masked pronouns. The usefulness of contrastive learning for pre-training BERT encoders has also been observed more generally, for instance for learning sentence, phrase and word embeddings \cite{gao-etal-2021-simcse,liu-etal-2021-fast,liu-etal-2021-mirrorwic,wang-etal-2021-phrase,li-etal-2022-uctopic}.

\paragraph{Leveraging Coreference Chains}
To the best of our knowledge, the idea of pre-training an entity encoder based on coreference chains has not yet been considered. However, a number of authors have proposed multi-task learning frameworks in which coreference resolution and entity typing are jointly learned, along with other tasks such as relation and event extraction \cite{luan-etal-2018-multi,wadden-etal-2019-entity}. Surprisingly, perhaps, such approaches have failed to outperform simpler entity typing (and relation extraction) models \cite{zhong-chen-2021-frustratingly}.

%% file: 3_our_approach.tex
% #==============================================================================================#
\section{Our Approach}
\label{sec:our_approach}
%In this section, we describe our proposed approach. 
In Section \ref{secBaseModel}, we first discuss the basic entity typing model that we rely on in this paper. Section \ref{subsec:pretraining_tasks} subsequently describes our proposed pre-training strategy based on coreference chains.

%************************************
\subsection{Entity Typing}\label{secBaseModel}
%We will rely on the method from \citet{zhong-chen-2021-frustratingly}, called PURE, as our base entity typing model.
Let us assume that we are given a sentence in which some entity mentions are highlighted, e.g.:
\begin{myquote}
\emph{[Alice] was unsure what was wrong with [the patient in front of her].}
\end{myquote}
\noindent Our aim is to assign (possibly fine-grained) semantic types to these entity mentions. For instance, using the FIGER \cite{DBLP:conf/aaai/LingW12} taxonomy, the first mention should be assigned the types \emph{Person} and \emph{Doctor}, while the second mention should be assigned \emph{Person}. 
To make such predictions, a given entity mention $e$ in sentence $s$ is first mapped to an embedding $\mathsf{Enc}(s,e)\in \mathbb{R}^n$ using an encoder. For the experiments in our paper, this encoder takes the form of a language model from the BERT family \cite{devlin-etal-2019-bert}. 
Specifically, we use the final-layer embedding of the head word of the given entity span as the representation of the mentioned entity. For instance, for the second mention in the aforementioned example, \emph{the patient in front of her}, we use the embedding of the head word, \emph{patient}, as the representation of the entity span.
%This is motivated by the fact that our entity encoder will be trained on the mention spans that are obtained from a coreference resolution system, which can be noisy. Rather than relying on exact span boundaries, we therefore only assume that the head word of each entity span can be correctly determined. 
This is motivated by the fact that the head word is most likely to reflect the semantic type of the entity \cite{choi-etal-2018-ultra}.  We find the head word using the SpaCy dependency 
parser\footnote{\url{https://spacy.io/api/dependencyparser}}. 

We pre-train the entity encoder $\mathsf{Enc}$ based on coreference chains, as will be explained in Section \ref{subsec:pretraining_tasks}. For each entity type $t$, we learn a vector $\mathbf{a}_t\in\mathbb{R}^n$ and bias term $b_t\in \mathbb{R}$. The probability that the mention $m$ should be assigned the type $t$ is then estimated as:
\begin{align}\label{eqLabelClassifier}
P(t | s,e) = \sigma(\mathbf{a}_t\cdot \mathsf{Enc}(s,e) + b_t)
\end{align}
with $\sigma$ the sigmoid function. This entity type classifier is trained using binary cross-entropy on a standard labelled training set. The encoder $\mathsf{Enc}$ is optionally also fine-tuned during this step. When using the classifier for entity typing, we assign all labels whose predicted probability is above 0.5.

%\blue{For the embedding of each token in the input sequence we predict its label from the label space using:
%\begin{align}\label{eqLabelClassifier}
%    P(t_i \mid \Tuple{\mathbf{e_1}, \dotsc, \mathbf{e_n}}, T)
%\end{align}
%where \textbf{$e_i$} is the token embedding given by our pre-trained entity encoder, $T$ is the label space, and
%$t_i$ is the predicted label.
%}

%Formally, we mask out the entity mention in the input sentence when fine-tuning entity typing. Then the sentence, ${s_j}, {=}, \Tuple{x_1, \dotsc, x_n}$, is fed into the pre-trained model, where $x_i$ is the $i$-th token in the sentence and $n$ is the total number of tokens in the sentence. Let  $\varepsilon$ denote a set of pre-defined entity types. For entity typing, we extract the vector representation of the masked out entity, $\bar{x_i}$, and predict its type, $y_e(\bar{x_i})  \in \varepsilon$. In principle, entity typing and FET are the same, except that entity typing is limited  to producing labels from a small set of entity classes, whereas FET predicts labels from a larger set.

%************************************
\subsection{Pre-training the Entity Encoder}
\label{subsec:pretraining_tasks}
To pre-train the entity encoder $\mathsf{Enc}$, we start from a collection of stories (e.g.\ news stories). Using off-the-shelf coreference resolution systems, we identify mentions within each story that are likely to refer to the same entity. Let us write $(s,e)$ to denote an entity mention $e$ appearing in sentence $s$. Then we consider the following self-supervision signal: if $(s_1,e_1)$ and $(s_2,e_2)$ are co-referring mentions, then the contextualised representations of $e_1$ and $e_2$ should be %similar.
close to each other in the embedding space.
In particular, we use a contrastive loss to encode that the representations of the tokens appearing in $e_1$ and $e_2$ should be more similar to each other than to the tokens appearing in the mentions of other entities.

Each mini-batch is constructed from a small set of stories $\{S_1,...,S_k\}$. Let us write $X_i$ for the set of entity mentions $(s,e)$ in story $S_i$ that belong to some coreference chain. %In other words, $(s,e)\in X_i$ if there is another entity mention $(s',e')$ in the same story that is linked to $(s,e)$ by the coreference resolution system. 
%For each $(s,e)\in X_i$ we let $C_{(s,e)}$ be the set of all mentions $(s',e')$  belonging to the same coreference chain. The mentions in $C_{(s,e)}$  will be treated as positive examples, i.e.\ we assume that their embeddings should be similar to that of $(s,e)$. 
To alleviate the impact of noisy coreference links, we adopt two strategies:
\begin{itemize}
\item We only include %$(s',e')$ in $C_{(s,e)}$ if this coreference link is 
coreference links that are predicted by two separate coreference resolution systems. This reduces the number of spurious links that are considered.
\item As negative examples, we only consider entity mentions from different stories. This prevents us from using entity mentions that refer to the same entity, but were missed by the coreference resolution system.
\end{itemize}
Let us write $T_i$ for the set of tokens of the mentions in $X_i$. For a given token $t$, we write $\mathsf{Enc}(t)$ for its contextualised representation. We write 
$T=T_1\cup ...\cup T_k$ and $T_{-i} = T\setminus T_i$. For a given token $t$, we write $C_t$ for the set of tokens that are part of the same coreference chain. The encoder is trained using InfoNCE \cite{DBLP:journals/corr/abs-1807-03748}:

\begin{align}\label{eqEntityEncoderLoss}
 %\sum_{i=1}^k\sum_{t\in T_i} \sum_{t'\in C_t}  \phi(t,t')
\sum_{i=1}^k\sum_{t\in T_i} \sum_{t'\in C_t}  \log\frac{\textit{exp}\Big(\frac{\cos(\mathsf{Enc}(t),\mathsf{Enc}(t'))}{\tau}\Big)}{\sum_{t''} \textit{exp}\Big(\frac{\cos(\mathsf{Enc}(t),\mathsf{Enc}(t''))}{\tau}\Big)}
\end{align}
where $t''$ in the denominator ranges over $T_{-i}\cup\{t\}$. The token pairs in the numerator correspond to positive examples, i.e.\ tokens whose embeddings should be similar, while the denominator ranges over both positive and negative examples.
%Note how mentions from the same document as $(s,e)$, but from a different coreference chain, are not used at all, as explained above. 
The temperature $\tau>0$ is a hyper-parameter, which controls how hard the separation between positive and negative examples should be. 

Given a mention $(s,e)$, the model can often infer the semantic type of the entity based on the mention span itself. To encourage the model to learn to identify cues in the sentence context, we sometimes mask the entity during training, following existing work on relation extraction \cite{baldini-soares-etal-2019-matching,peng-etal-2020-learning}. Specifically, for each input $(s,e)\in X$, with 15\% probability we replace the head of the entity span by the [MASK] token. Note that, unlike previous work, we only mask the head word of the phrase. %We find the head word using the SpaCy dependency parser\footnote{\url{https://spacy.io/api/dependencyparser}}.

Finally, following \citet{baldini-soares-etal-2019-matching}, we also use the Masked Language Modelling objective during training, to prevent catastrophic forgetting. Our overall loss thus becomes:
$$
\mathcal{L} = \mathcal{L}_{\text{entity}} +  \mathcal{L}_{\text{MLM}}
$$
where $\mathcal{L}_{\text{entity}}$ is the loss function defined in \eqref{eqEntityEncoderLoss} and $\mathcal{L}_{\text{MLM}}$ is the masked language modelling objective from BERT \cite{devlin-etal-2019-bert}.
% Some discussion

%In contrast to~\ac{BERT}, where every type of token in the inputs is randomly masked, we systematically only mask out a percentage of entity mentions in the inputs and let the model predict the masked entity mentions. We represent the \ac{MLM} loss as $L_{MLM}$.

%% file: 4_experiments_and_results.tex
% #==============================================================================================#
\section{Experimental Analysis}
\label{sec:experiments_and_results}
In this section, we evaluate the performance of our proposed strategy on (fine-grained) entity typing.\footnote{Our implementation and pre-trained models are available at \url{https://github.com/fmtumbuka/EACL_EnCore}}

\paragraph{Experimental Setup}
In all our experiments, we initialise the entity encoder with a pre-trained language model.
We consider 
{\texttt{bert-base-uncased}}\footnote{\url{https://huggingface.co/docs/transformers/model_doc/bert}}, {\texttt{albert-xxlarge-v1}}\footnote{\url{https://huggingface.co/docs/transformers/model_doc/albert}} 
and {\texttt{roberta-large}}\footnote{\url{https://huggingface.co/docs/transformers/model_doc/roberta}} for this purpose, as these are commonly used for entity typing. %The entity encoders are then fine-tuned based on coreference chains, as explained in Section \ref{subsec:pretraining_tasks}. 
We use the Gigaword corpus\footnote{\url{https://catalog.ldc.upenn.edu/LDC2003T05}} as the collection of stories. This corpus consists of around 4 million news stories from four different sources. We use two state-of-the-art coreference resolution systems: the \textbf{Explosion AI} system Coreferee v1.3.1\footnote{\url{https://github.com/explosion/coreferee}} and the \textbf{AllenNLP} coreference model\footnote{\url{https://demo.allennlp.org/coreference-resolution}}.
%\item[Explosion AI] The Explosion AI system for resolving coreferences, Coreferee v1.3.1\footnote{\url{https://github.com/explosion/coreferee}}, reads each document naturally from left to right while constructing chains of anaphors and independent noun phrases. For each anaphor, the neural ensemble's highest-scoring interpretation is preferred. However, in some cases, this interpretation might not be allowed because it would result in a semantically inconsistent chain. In cases where the highest-scoring interpretation is not allowed, the next highest-scoring interpretation is then tried. Coreferee builds on SpaCy's  Neuralcoref\footnote{\url{https://github.com/huggingface/neuralcoref}}.
%\item[AllenNLP] For each span in the document, the AllenNLP coreference model\footnote{\url{https://demo.allennlp.org/coreference-resolution}} obtains an embedded representation. The model then scores the span representations and filters out spans that are unlikely to occur in a coreference cluster. The model determines which antecedent span is coreferent with the remaining spans. The AllenNLP coreference system extends the system proposed by~\citet{lee-etal-2018-higher}. Unlike the original model, which used GloVe embeddings, AllenNLP's system obtains span representations from SpanBERT~\cite{joshi-etal-2020-spanbert}.
%\end{description}
As explained in Section \ref{subsec:pretraining_tasks}, we only keep coreference links that are predicted by both of these systems. 
%\st{Once pre-trained, we freeze the entity encoder and train an entity type classifier on the standard training set for each benchmark. We will refer to our model as \emph{EnCore}.}
Once the encoder has been pre-trained, we train an entity type classifier on the standard training set for each benchmark. We report results for two different variants of this process: one where the entity encoder is fine-tuned while training the entity type classifiers and one where the encoder is frozen.
We will refer to these variants as \emph{EnCore} and \emph{EnCore-frozen}, respectively.
We train all of the models for 25 epochs with the AdamW optimizer \cite{loshchilov2018decoupled} and save the checkpoint with the best result on the validation set. The temperature $\tau$ in the contrastive loss was set to 0.05. 

%\frank{For multi-label entity typing, we consider predictions above a threshold of 0.5 as true.}
%We were unable to conduct grid search for every single experiment separately because of the large number of experiments that have been carried out. Instead, we chose a set of hyperparameters that worked well across a large number of initial training runs and kept them constant across all experiments.

%**********************************
\begin{table}
\footnotesize
\centering
\begin{tabular}{lcccc}
\toprule
\textbf{Dataset} & \textbf{\# Types} & \textbf{Train} & \textbf{Dev.} & \textbf{Test}\\
\midrule
ACE 2005 & 7 & 26.5K & 6.4K & 5.5K \\
OntoNotes & 89 & 3.4M & 8K & 2K\\
FIGER & 113 & 2M & 1K & 0.5K\\
\bottomrule
\end{tabular}
\caption{Overview of the considered benchmarks, showing the number of entity types, and the number of entity mentions in the training, development and test sets. \label{tabOverviewDatasets}}
\end{table}

\paragraph{Benchmarks}
Our central hypothesis is that the proposed pre-training task makes it possible to learn finer-grained entity representations. As such, we focus on fine-grained entity typing as our main evaluation task. We use the OntoNotes \cite{DBLP:journals/corr/GillickLGKH14} and FIGER \cite{DBLP:conf/aaai/LingW12} benchmarks. OntoNotes is based on the news stories from the OntoNotes 5.0 corpus\footnote{\url{https://catalog.ldc.upenn.edu/LDC2013T19}}. We use the entity annotations that were introduced by \citet{DBLP:journals/corr/GillickLGKH14}, considering a total of 89 different entity types (i.e.\ 88 types + \emph{other}). They also introduced a distantly supervised training set, consisting of 133K automatically labelled news stories. FIGER considers a total of 113 types (i.e.\ 112 types + \emph{other}). The test set consists of sentences from a student newspaper from the University of Washington, two local newspapers, and two specialised magazines (on photography and veterinary). Along with this test set, they also provided automatically labelled Wikipedia articles for training. For fine-grained entity typing, we report the results in terms of macro and micro-averaged F1, following the convention for these benchmarks. 

We also experiment on standard entity typing, using the ACE 2005 corpus\footnote{\url{https://catalog.ldc.upenn.edu/LDC2006T06}}, which covers the following text genres: broadcast conversation, broadcast news, newsgroups, telephone conversations and weblogs. It differentiates between 7 entity types. For this benchmark, the entity spans are not provided. We thus need to identify entity mentions in addition to predicting the corresponding types. We treat the problem of identifying entity span as a sequence labelling problem.  We follow the strategy from \citet{hohenecker-etal-2020-systematic}, but start from our pre-trained entity encoder rather than a standard LM. We summarise this strategy in Appendix \ref{secSpanDetection}.
We use the standard training/development/test splits that were introduced by \citet{li-ji-2014-incremental}. Following standard practice, we report the results in terms of micro-averaged F1. We take individual sentences as input. Existing work on this benchmark jointly evaluates span detection and entity typing, i.e.\ a prediction is only correct if both the span and the predicted type are correct. We will refer to this as the \emph{strict} evaluation setting, following \citet{bekoulis-etal-2018-adversarial}. We also consider the \emph{lenient} setting from, where a prediction is scored as correct as soon as the type is correct and the predicted span \emph{overlaps} with the gold span.

%\blue{
%In this study, we employ two evaluation schemes as defined in~\cite{bekoulis-etal-2018-adversarial}. The first scheme is the \textbf{strict} scheme, in which the entity is scored as correct if both the entity boundaries and entity types are correct. The second approach is the \textbf{lenient} scheme, in which the entity is scored as correct if at least one of its tokens is assigned the correct type. These evaluation schemes allow us to accurately compare our work to existing work on the three benchmark datasets that we consider.
%}
%While this means that we cannot compare our results with published work, we can still analyse the impact of our pre-training strategy. In particular, we want to use this benchmark to find out whether our approach has any benefit for coarse-grained entity types, or whether its benefit is primarily in improving the discrimination between fine-grained types. We have also used this benchmark for development purposes, e.g.\ for deciding how to represent entity spans and which masking strategy to apply. 

% ACE 2005
% newswire + online forums
% 511 docs
% 14.5K sentences
% 38.3k entities
% 7 entity types

% Splits from \citet{li-ji-2014-incremental} into 351 for training, 80 for development, and 80 for testing
% TRAIN: 351 documents, 7273 sentence, 26470 mentions
% DEVELOPMENT: 80 documents, 1765 sentences, 6421 mentions
% TEST: 80 documents, 1535 sentences, 5476 mentions

% Splits used in PURE: sentences: 10051 training, 2424 development, 2050 testing

Table \ref{tabOverviewDatasets} summarises the main characteristics of the considered datasets.

%*****************************
\paragraph{Baselines}
%For ACE 2005, as already mentioned, we cannot directly compare our results with published work, since we assume that gold spans are given. Our focus for this dataset will therefore be on comparing our full model with a variant in which the entity encoder is learned in the standard way, i.e.\ by fine-tuning the language model on the distantly supervised training set while training the label classifiers. We will refer to this version as the \emph{base} model. 
We report results for a number of simplified variants of our main model. First, we consider a variant which uses the same strategy for training the entity type classifier as our full model, but without pre-training the entity encoder on the Gigagword corpus. This variant is referred to as the \emph{base model}. Second, we investigate a setup in which the entity encoder is pre-trained on Gigaword, but only using the \ac{MLM} objective. This setting, which we refer to as \emph{MLM-only}, allows us to analyse to what extent improvements over the base model are due to the continued training of the language model.
%In addition, we consider PURE~\cite{zhong-chen-2021-frustratingly} to get the entity spans in an input sentence, and use \emph{EnCore} to get the entity types. 
%We refer to this setting in the results section as PURE-EnCore.
%Note that \steven{both the aforementioned} setups and our full model (EnCore) follow the approach described in Section \ref{secBaseModel}. Comparing \steven{these} models will thus allow us to directly assess the impact of our proposed strategy for training the entity encoder. 
%For EnCore, we include a variant where the entity encoder is frozen when training the entity type classifier; this variant is referred to as \emph{EnCore-frozen}.
%}

For reference, we also compare our models with the published results of state-of-the-art models.
For fine-grained entity typing, we consider the following baselines:
\textbf{DSAM} \cite{DBLP:journals/access/HuQXP21} is an LSTM-based model, which we include as a competitive baseline;
\textbf{Box4Types} \cite{onoe-etal-2021-modeling} uses hyperboxes to represent mentions and types, to take advantage of the hierarchical structure of the label space;
\textbf{PICOT} \cite{zuo-etal-2022-type} uses a contrastive learning strategy based on the given type hierarchy;
\textbf{Relational Inductive Bias (RIB)} \cite{DBLP:conf/ijcai/Li0WL21} uses a graph neural network to model correlations between the different labels. Entity mentions are encoded using a transformer layer on top of pre-trained ELMo \cite{peters-etal-2018-deep} embeddings;
\textbf{LITE} \cite{li-etal-2022-ultra} assigns entity types by fine-tuning a pre-trained Natural Language Inference model;
\textbf{SEPREM} \cite{xu-etal-2021-syntax} improves on the standard RoBERTa model by exploiting syntax during both pre-training and fine-tuning, and then using a standard entity typing model on top of their pre-trained model;
\textbf{MLMET} \cite{dai-etal-2021-ultra} extends the standard distantly supervised training data, using the BERT masked language model for generating weak labels;
\textbf{DenoiseFET} \cite{DBLP:conf/ijcai/Pan0022} uses a denoising strategy to improve the quality of the standard distantly supervised training set, and furthermore exploits prior knowledge about the labels, which is extracted from the parameters of the decoder of the pre-trained BERT model; \textbf{PKL} \cite{DBLP:journals/corr/abs-2305-12802} improves on DenoiseFET by incorporating pre-trained label embeddings.
%LITE, MLMET and DenoiseFET all use prior knowledge about the type labels, which is distilled in different ways from a language model. In contrast, our model treats the entity types as abstract labels, which puts it at a disadvantage for rare types in particular. 

For ACE 2005, we consider the following baselines:
\textbf{DyGIE++} \cite{wadden-etal-2019-entity} uses multi-task learning to jointly train their system for coreference resolution, entity typing, relation extraction and event extraction;
\textbf{TableSeq} \cite{wang-lu-2020-two} jointly trains a sequence encoder for entity extraction and a table encoder for relation extraction;
\textbf{UniRe} \cite{wang-etal-2021-unire} also uses a table based representation, which is shared for entity and relation extraction;
\textbf{PURE} \cite{zhong-chen-2021-frustratingly} uses BERT-based models to get contextualised representations of mention spans, which are fed through a feedforward network to predict entity types;
\textbf{PL-Marker} \cite{ye-etal-2022-packed} builds on PURE by introducing a novel span representation.
%Our model is most closely related to PURE. Apart from the fact that we train our entity encoder on coreference chains, the main difference is that PURE uses the embedding of the first and last token of the entity span, whereas we use special tokens to obtain span encodings. They also use a feedforward network to predict the label types, whereas we use a simple classifier, i.e.\ Eq.\ \eqref{eqLabelClassifier}. 

%***************************************
\subsection{Results}
Table~\ref{tab:comparisons_to_baselines_FET} summarises the results for fine-grained entity typing. As can be seen, EnCore outperforms the base and \textit{MLM-only} models by a large margin, which clearly shows the effectiveness of the proposed pre-training task. Remarkably, EnCore-frozen performs only slightly worse. The best results are obtained with {\texttt{roberta-large}}. Our model furthermore outperforms the baselines on both OntoNotes and FIGER, except that RIB achieves a slightly higher micro-averaged F1 on FIGER.  It should be noted that several of the baselines introduce techniques that are orthogonal to our contribution in this paper, e.g.\  denoising the distantly supervised training sets (DenoiseFET), incorporating prior knowledge about the type labels (PKL) and exploiting label correlations (RIB), which would likely bring further benefits when combined with our pre-training strategy.

%For FIGER, {\texttt{bert-base-uncased}} does not seem to be sufficiently expressive to allow for competitive results. 
%We can see that our full model (EnCore) consistently outperforms the MLM-only variant, which in turn consistently outperforms the Base model. Moreover, EnCore still performs well even when the encoder is frozen.
% note that the two baselines which rely on {\texttt{bert-base} did not report any results for FIGER. One important distinction between both benchmarks is that the ground truth for OntoNotes only includes labels that can be inferred from the given sentence \cite{DBLP:journals/corr/GillickLGKH14}. This explains why models that explicitly capture label correlations, such as RIB, perform comparatively better on FIGER. It also means that models need more prior knowledge about named entities  to do well on FIGER, which explains why {\texttt{bert-base}} performs comparatively worse on this benchmark.

Table~\ref{tab:comparisons_to_baselines_ET} summarises the results for standard entity typing (ACE 2005). 
%\steven{We show results for two settings. The \emph{strict} evaluation is the standard setting where both the entity spans and the corresponding types have to be predicted. In this setting, a prediction is only correct if both the span and the type are correct. For our models, we use the entity boundaries predicted by PURE. In the \emph{lenient} evaluation, we provide the model with the gold entity spans and only evaluate whether the predicted types are correct.}
We can again see that EnCore consistently outperforms the MLM-baseline, which in turn consistently outperforms the base model.
Comparing the different encoders, the best results for our full model are obtained with {\texttt{albert-xxlarge-v1}}, which is consistent with what was found in previous work \cite{zhong-chen-2021-frustratingly,ye-etal-2022-packed}. 
Finally, we can see that our full model outperforms all baselines.

{
\begin{table}[t]
    \footnotesize
        \centering
        \begin{tabular}{llCCCC}
            \toprule
            \textbf{Model}       &             
            \textbf{LM}       & 
            \multicolumn{2}{c}{\textbf{OntoNotes}}               &
            \multicolumn{2}{c}{\textbf{FIGER}}                  \\ 
            \cmidrule(lr){3-4}\cmidrule(lr){5-6}
            &
            &
            macro &
            micro &
            macro &
            micro \\
            \midrule
            DSAM        &      LSTM  & 83.1         & 78.2          & 83.3         & 81.5      \\
            Box4Types   &      BL    & 77.3         & 70.9          & 79.4         & 75.0      \\
            PICOT       &     BL    & 78.7         & 72.1          & 84.7         & 79.6      \\
            RIB         &    ELMo  & 84.5         & 79.2          & 87.7         & \textbf{84.4}      \\
            LITE        &    RL    & 86.4         & 80.9          & 86.7         & 83.3      \\
            SEPREM      &     RL    & -            & -             & 86.1         & 82.1      \\
            MLMET       &     BBc   & 85.4         & 80.4          & -            & -         \\
            %DenoiseFET  &      BBc   & 87.1         & 81.5          & -            & -         \\  
            DenoiseFET & BB & 87.2 & 81.4 & 86.2 & 82.8 \\
            DenoiseFET & RL & 87.6 & 81.8 & 86.7 & 83.0 \\
            PKL         &  BB & 87.7 & 81.9 & 86.8 & 82.9 \\
            PKL         &  RL & 87.9 & 82.3 & 87.1 & 83.1 \\
            \midrule
              %                      &       &  BB     & 74.9	     & 71.3	         & 77.8	        & 74.6     \\ 
  %                      &   Yes &  ALB    & 76.1	     & 73.5	         & 78.3	        & 77.4    \\
  %                      &       &  RL     & 79.9	     & 76.4	         & 81.2	        & 78.7     \\ %\cline{2-7}
                          &  BB     & 76.9	   & 72.9	       & 78.6	        & 76.1     \\ 
            Base model    &  ALB    & 77.9	     & 74.8	         & 80.2	          & 77.4     \\  
                          &  RL     & 82.8	   & 80.1	       & 82.3	        & 79.5     \\ 
            \midrule
                       % &       &  BB     & 78.1	     & 76.9	         & 79.5	        & 76.9     \\ 
                       % &   Yes &  ALB    & 81.3	     & 80.6	         & 80.8	        & 79.5     \\  
                       % &       &  RL     & 84.9	     & 80.1	         & 85.6	        & 82.3     \\ %\cline{2-7}
                         &  BB     & 81.6	     & 78.7	         & 80.2	        & 77.9     \\ 
            MLM-only     &  ALB    & 82.7	      & 79.8	      & 81.5	     & 79.6     \\  
                         &  RL     & 85.4	     & 81.4	         & 85.8	        & 82.1     \\             
            \midrule
                            &   BB     & 87.3	     & 80.6       & 87.1	        & 82.2      \\ 
            EnCore-frozen   &   ALB    & 87.9        & 81.9       & 87.7	        & 82.9      \\ 
                            &   RL     & 88.3        & 82.7	      & 87.8	        & 83.6     \\
            \midrule
                        &  BB     & 87.6	      & 81.9	        & 87.3	        & 82.9      \\ 
            EnCore      &  ALB    & 88.7	      & 82.9            & 87.9	        & 83.8      \\ 
                        &  RL     &\textbf{88.9}  & \textbf{83.4}	&\textbf{88.4}	& 84.1 \\
            \bottomrule
        \end{tabular}
        \caption{%
       Results for fine-grained entity typing, in terms of macro-F1 and micro-F1 (\%). BB stands for {\texttt bert-base-uncased}, BBc stands for {\texttt bert-base-cased}, BL stands for {\texttt bert-large-uncased},  ALB stands for {\texttt{albert-xxlarge}}
        and RL stands for {\texttt{roberta-large}} DenoiseFET results are taken from \cite{DBLP:journals/corr/abs-2305-12802}; all other baseline results are taken from the original papers.
    	} \label{tab:comparisons_to_baselines_FET}%
\end{table}

% ################ Previous Table 2 #########################
 \begin{table}[t]
 	\begin{center}
 		\footnotesize{
 			\centering{ %\setlength\tabcolsep{3pt}
 	    \begin{tabular}{lDDDDDD}
 		\toprule
        & \multicolumn{3}{c}{\textbf{Strict}} & \multicolumn{3}{c}{\textbf{Lenient}}\\
 	\cmidrule(lr){2-4}\cmidrule(lr){5-7}	
    &\textbf{BB}  & \textbf{ALB}   & \textbf{RL} &\textbf{BB}  & \textbf{ALB}   & \textbf{RL}
         \\
   \midrule
DyGIE++$^\diamond$ & 88.6 & - & - & -& -& -\\
UniRe$^\diamond$ & 88.8 & 90.2 & -& -& -& -\\
PURE$^\diamond$ & 90.1  & 90.9  & - & -& -& -\\
PL-Marker$^\diamond$ & 89.8 & 91.1 & - & -& -& -\\
\midrule
PURE   & 88.7 & 89.7 & - & -& -& -\\
TableSeq & - & 89.4 & 88.9 & - & - & -\\
\midrule
Base model    & 86.8 & 87.1 & 86.9  & 90.3 & 90.8 & 90.6\\
MLM-only      & 87.1 & 87.8 & 87.5  & 90.7 & 91.2 & 90.9\\
EnCore-frozen & 89.9 & 90.5 & 90.1  & 91.8 & 92.3 & 92.0\\
EnCore        & \textbf{90.8} & \textbf{91.9} & \textbf{91.0}  & \textbf{92.4} & \textbf{93.1} & \textbf{92.6}\\
\bottomrule                                
 		\end{tabular}}}
 	\caption{%
  Results for entity typing on ACE 2005, in terms of micro-F1 (\%). BB stands for {\texttt bert-base-uncased}, ALB stands for {\texttt{albert-xxlarge}}
     and RL stands for {\texttt{roberta-large}. Configurations with $^\diamond$ rely on cross-sentence context and are thus not directly comparable with our method.}
 	} \label{tab:comparisons_to_baselines_ET}%
 	\end{center}
 \end{table}

%***************************************
\subsection{Analysis}\label{secAnalysis}

\begin{table}[t]
\label{table:pretraining_and_finetuning_strategy}
	\begin{center}
		\footnotesize{
			\centering{\setlength\tabcolsep{4pt}        
	    \begin{tabular}{lcc}
		\toprule
		%\multirow{2}{*}{\textbf{Strategy}}
  \textbf{Strategy}&
		%\multicolumn{1}{c}{\textbf{ACE05}}                         %    &
        \multicolumn{2}{c}{\textbf{OntoNotes}}                        \\
        %\cmidrule(lr){2-2}
        \cmidrule(lr){2-3}
        %& micro 
        & macro & micro\\
%		\multicolumn{1}{c}{}                                             &
%		\multicolumn{2}{c}{\textbf{macro-F1}}                            \\ 
		\midrule

        % \multicolumn{2}{c}{PURE (Bert-base) - paper}                      &
        % \multicolumn{1}{c}{88.71}                                         &
        % \multicolumn{1}{c}{-}                                             \\

        % \multicolumn{2}{c}{PURE (Bert-base) - Ours}                        &
        % \multicolumn{1}{c}{86.93}                                          &
        % \multicolumn{1}{c}{-}                                              \\
        
        \multicolumn{1}{l}{MASK}                         &
        % entity is masked, get mask token representation
      %  \multicolumn{1}{c}{79.8}                                   %       &
        \multicolumn{1}{c}{70.7}                                          &
        \multicolumn{1}{c}{66.8}                                          \\

        \multicolumn{1}{l}{Prompt}                                        &
        % Sentence <SEP> ????
      %  \multicolumn{1}{c}{77.9}                                  %        &
        \multicolumn{1}{c}{72.1}                                          & 
        \multicolumn{1}{c}{68.7}                                          \\

     %   \multicolumn{1}{l}{Multiple Prompts}                   &
     %   \multicolumn{1}{c}{75.11}                                          &
     %   \multicolumn{1}{c}{79.43}                                          %\\
        
        \multicolumn{1}{l}{Masked triple}                                 &
        % Sentence <SEP> ????     
      %  \multicolumn{1}{c}{78.4}                                  %        &
        \multicolumn{1}{c}{72.8}                                          & 
        \multicolumn{1}{c}{69.4}                                         \\

        \multicolumn{1}{l}{Special tokens: full span}                     &
       % \multicolumn{1}{c}{89.1}                                  %        &
        \multicolumn{1}{c}{75.2}                                          & 
        \multicolumn{1}{c}{70.8}                                          \\

        \multicolumn{1}{l}{Special tokens: head}                         &
      %  \multicolumn{1}{c}{89.6}                                   %      &
        \multicolumn{1}{c}{76.1}                                         & 
        \multicolumn{1}{c}{71.3}                                         \\

        \midrule
        %\multicolumn{1}{l}{Entity mask tensor}                    
        \multicolumn{1}{l}{Head word}                    
        &
       % \multicolumn{1}{c}{90.3}                                   %      &
        \multicolumn{1}{c}{76.9}                                         & 
        \multicolumn{1}{c}{72.9}                                         \\
        
		\bottomrule
		\end{tabular}}}
	\end{center} 
	\caption{Comparison of different strategies for encoding entity spans (using {\texttt{bert-base-uncased}}). \label{tabAnalysisSpans}
	} 
\end{table}

\begin{table}[t]
	\begin{center}
		\footnotesize{
			\centering{\setlength\tabcolsep{4pt}
        %\resizebox{\textwidth}{!}{
	    \begin{tabular}{llccc}
		\toprule
		\textbf{Neg.\ samples} & \textbf{Masking} &
		\textbf{ACE05}                                       &
        \multicolumn{2}{c}{\textbf{OntoNotes}}                                  \\
       \cmidrule(lr){4-5}
       && micro & macro & micro\\
  
        \midrule
        Same story & None & 83.9 & 82.1 & 74.9\\
        Same story & Entire span & 84.7 & 82.9 & 75.3\\
        Different stories & Entire span & 88.8 & 86.2 & 78.9\\
       
        \midrule
        
        Different stories & Head & 91.8 & 87.3 & 80.6 \\
        
		\bottomrule
		\end{tabular}}}
 %   }
	\end{center} 
	\caption{Comparison of different strategies for pre-training the entity encoder (using {\texttt{bert-base-uncased}}). \label{tabAnalysisPretraining}
	} 
\end{table}

\begin{table}[t]
\label{table:coreference_systems}
	\begin{center}
		\footnotesize{
			\centering{\setlength\tabcolsep{4pt}        
	    \begin{tabular}{lccc}
		\toprule
		\textbf{Coreference Systems}                        &
		\multicolumn{1}{c}{\textbf{ACE05}}                                     &
        \multicolumn{2}{c}{\textbf{OntoNotes}}                                \\%& \multicolumn{1}{c}{\textbf{FIGER}}                                       \\
     %   \cline{3-5}
%		\multicolumn{2}{c}{}                                                     &
%		\multicolumn{3}{c}{\textbf{$F_1$}}                                        \\ 
     \cmidrule(lr){3-4}
                                    & micro    & macro    & micro   \\
		\midrule
        Explosion AI                & 86.4     & 83.4     & 79.4    \\ 
        AllenNLP                    & 90.7     & 86.8     & 80.1    \\ 
        Explosion AI + AllenNLP     & 91.8     & 87.3    & 80.6      \\  
		\bottomrule
		\end{tabular}}}    
	\end{center} 
	\caption{
    Comparison of different coreference resolution strategies (using {\texttt{bert-base-uncased}}). \label{tabComparisonCoref}} 
\end{table}
We now analyse the performance of our method in more detail. For this analysis, we will focus on ACE 2005 under the lenient setting and OntoNotes.
Throughout this section, unless mentioned otherwise, we use {\texttt{bert-base-uncased}} for the encoder.

\paragraph{Encoding Entity Spans}
We represent entities using the embedding of the head word.
In Table \ref{tabAnalysisSpans} we compare this approach with the following alternatives:
\begin{description}
\item[MASK] We replace the entity mention by a single MASK token and use the final-layer encoding of this token as the embedding of the entity. 
\item[Prompt] Given a mention $(s,e)$, we append the phrase ``The type of $e$ is [MASK].'' The final-layer encoding of the MASK-token is then used as the mention embedding.
\item[Masked triple] This strategy is similar to \emph{Prompt} but instead of appending a sentence, we append the phrase ``<$e$, hasType, [MASK]>''.
\item[Special tokens: full span] We add the special tokens \textit{<m>} and \textit{</m>} around the entire entity span. We take the final-layer encoding of the \textit{<m>} token as the embedding of the entity. 
\item[Special tokens: head] In this variant, we add the special tokens \textit{<m>} and \textit{</m>} around the head word of the entity span. %We take the final-layer encoding of the \textit{<m>} token as the embedding of the entity mention.
%\item[Entity mask tensor] This is the method adopted by our model. In this case, for each input sentence, we come up with an according tensor of 0s and 1s where 1s to represent tokens that belong to entity spans and 0s to represent non-entity span tokens. To get the entity mention embeddings, we only retrieve embeddings where the entity mask tensor has 1s.
\item[Head word] This is the method adopted in our main experiments. In this case, we simply use the embedding of the head word of the entity mention, without using special tokens.
\end{description}
In all cases, we use the entity typing model that was described in \ref{secBaseModel}. 
Note that we do not consider ACE 2005 for this analysis, as the entity spans have to be predicted by the model for this dataset, which means that aforementioned alternatives cannot be used.
For this analysis, we train the entity encoder on the training data of the considered benchmark, without using our coreference based pre-training strategy. The results in Table \ref{tabAnalysisSpans} show that using the embedding of the head word clearly outperforms the considered alternatives. Another interesting observation is that encapsulating the head of the entity mention performs slightly better than encapsulating the entire entity span, whereas it is the latter variant that is normally used in the literature. It is also notable, and somewhat surprising, that \emph{Masked triple} outperforms \emph{Prompt}.

\paragraph{Pre-training Strategies}
In Table \ref{tabAnalysisPretraining} we compare four strategies for pre-training the entity encoder based on coreference chains.  In particular, we analyse the effect of two aspects:
\begin{itemize}
\item When training our model, the negative examples for the contrastive loss (Section \ref{subsec:pretraining_tasks}) are always selected from other stories. Here we analyse the impact of choosing these negative examples from the same story instead.
\item During training, in 15\% of the cases, we mask the head of the entity span. Here we consider two other possibilities: (i) never masking the entity span and (ii) masking the entire span.
\end{itemize}
Choosing the negative examples from the same story has a number of implications. First, it may mean that false negatives are included (i.e.\ coreference links that were missed by the system). Second, it means that the overall number of negative examples becomes smaller, since they have to come from a single story. However, these downsides may be offset by the fact that negative examples from the same story may be harder to discriminate from the positive examples, since the story context is the same, and using harder negatives is typically beneficial for contrastive learning.
%\frank{All of the analysis in this section is based on the best strategy for encoding entity spans in Table~\ref{tabAnalysisSpans} and only considers coreference chains if both coreference resolution systems agree.}
For this analysis we use EnCore-frozen.
As can be seen in Table \ref{tabAnalysisPretraining}, choosing negative examples from the same story overall has a clearly detrimental impact. We also find that masking is important, where masking only the head of the entity span leads to the best results. This masking strategy has not yet been used in the literature, to the best of our knowledge.

%***************
\paragraph{Coreference Resolution}
In Table \ref{tabComparisonCoref} we analyse the importance of using only high-quality coreference links. 
%\blue{The most effective entity span encoding method from Table~\ref{tabAnalysisSpans} and the most effective pre-training method from Table~\ref{tabAnalysisPretraining} are combined here.}
In particular, we compare three configurations: (i) using all links predicted by the Explosion AI system; (ii) using all links predicted by the AllenNLP system; and (iii) using only the links that are predicted by both systems. For this analysis, we use EnCore-frozen. As can be seen, the AllenNLP system overall performs better than the Explosion AI system. However, the best results are obtained by combining both systems.

\paragraph{Performance on Fine and Coarse Labels}
In Table~\ref{tabComparisonLabelOrders} we compare our full model with the \textit{MLM-only} variant on different partitions of the OntoNotes test set. 
% We specifically examine how EnCore compares to \textit{MLM-only} on i) samples with a single label (5.3K), ii) samples with two labels (3.0K), and iii) samples with three labels (0.6K). Examples with one label only require the model to identify the top-level entity type (e.g.\ {\texttt{/organisation}}), whereas examples with two labels require making finer-grained distinctions (e.g.\  {\texttt{/organisation}} and {\texttt{/organisation/company}}), and examples with three labels involve a further refinement (e.g.\ {\texttt{/organisation}}, {\texttt{/organisation/company}}  and {\texttt{/organization/company/broadcast}}). As can be seen, EnCore outperforms \textit{MLM-only} in all cases, but the difference is smallest in the one-label case. 
We specifically compare EnCore and \textit{MLM-only} on those examples with one-level labels (5.3K);  two-level labels (3.0K); and three-level labels (0.6K). Examples with one-level labels only require the model to determine the top-level entity type (e.g\ {\texttt{/organisation}}). Examples with two-level labels call for more precise finer-grained differentiations (e.g.\  {\texttt{/organisation}} and {\texttt{/organisation/company}}). Examples with three-level labels call for even more precision (e.g.\ {\texttt{/organisation}}, {\texttt{/organisation/company}}  and {\texttt{/organization/company/broadcast}}). EnCore performs better than \textit{MLM-only} in every scenario, as can be observed, with the difference being least pronounced in the case of single-level labels.
This supports the idea that our pre-training technique is particularly useful for learning finer-grained entity types. A more detailed breakdown of the results, which is provided in the appendix, shows that EnCore consistenly outperforms \textit{MLM-only} on all labels, both for OntoNotes and FIGER. 
%In Appendix~\ref{app:modelOutputComparison}, we go into additional detail about how EnCore provides more confident predictions than MLM-only models.}

\begin{table}[t]
	\begin{center}
		\footnotesize{
			\centering{\setlength\tabcolsep{3pt}        
	    \begin{tabular}{lcccccc}
		\toprule
		\textbf{Model}                                                        &
		\multicolumn{2}{c}{\textbf{One Label}}                                &
        \multicolumn{2}{c}{\textbf{Two Labels}}                               &
        \multicolumn{2}{c}{\textbf{Three labels}}                             \\
     \cmidrule(lr){2-3}
     \cmidrule(lr){4-5}
     \cmidrule(lr){6-7}
                    & macro     & micro        & macro    & micro       & macro    & micro       \\
		\midrule
        MLM-only    & 79.8      & 75.6         & 53.0     & 50.9        &  39.1    & 38.4        \\
        EnCore      & 82.7      & 78.7         & 59.8     & 58.5        &  44.6    & 43.6        \\
		\bottomrule
		\end{tabular}}}    
	\end{center} 
	\caption{
    Comparison of the MLM-only and EnCore models (using {\texttt{roberta-large}}) on partitions of the OntoNotes test set. \label{tabComparisonLabelOrders}} 
\end{table}

%% file: 5_conclusion.tex
% #==============================================================================================#
\section{Conclusion}\label{sec:conclusion}
We have proposed a strategy which uses coreference chains to pre-train an entity encoder. Our strategy relies on the natural idea that coreferring entity mentions should be represented using similar vectors. Using a contrastive loss for implementing this intuition, we found that the resulting encoders are highly suitable for (fine-grained) entity typing. %In particular, by training a simple multi-label classifier on top of the frozen mention encoder, we were able to outperform the state-of-the-art in two standard benchmarks for fine-grained entity typing. 
% In our analysis, we found that restricting our strategy to high-quality coreference links was important for its success. We also found that focusing on the head of the entity span, rather than the span itself, was beneficial, both when it comes to representing the entity span with special tokens (where encapsulating only the head worked better) and when it comes to masking entities during training (where only masking the head was most helpful).
In our analysis, we found that restricting our strategy to high-quality coreference links was important for its success. We also found that focusing on the head of the entity span, rather than the span itself, was beneficial, both when it comes to representing the entity span and when it comes to masking entities during training (where only masking the head was found to be most helpful).
%In addition, we also found that our pre-training strategy helped in learning finer-grained entity types as our model showed much better performance in multi-label settings than counterparts.

% Main idea
% Frozen entity encoder prevents over-fitting
% Only considered a simple entity typing model here; in future work we can combine this with pre-trained label embeddings for UFET.
% Other avenue for future work is to explore the usefulness of coreference chains for pre-training relation extraction systems.

%% file: 6_limitations.tex
% #==============================================================================================#
\section{Limitations}

Our model is pre-trained on individual sentences. This means that during testing, we cannot exploit cross-sentence context. Prior work has found such cross-sentence context to be helpful for benchmarks such as ACE2005, so it would be of interest to extend our model along these lines. Furthermore, we have not yet applied our model to ultra-fine entity typing, as this task requires us to cope with labels for which we have no, or only very few training examples. This would require combining our entity encoder with entity typing models that can exploit label embeddings, such as UNIST \cite{huang-etal-2022-unified}, which we have left as an avenue for future work.

%% file: A_Appendix.tex
\section{Entity Span Detection}\label{secSpanDetection}
 We treat the problem of entity span detection as a sequence labelling problem, following the strategy from \citet{hohenecker-etal-2020-systematic}. Specifically, each token in the input sentence is then labelled with an appropriate tag, which could either be one of the entity types from the considered dataset or a tag which denotes that the token does not belong to any entity span. To assign these tags, we again use the encoder that was pre-trained on coreference chains. However, rather than looking only at the head word of a given entity span, we now consider the embedding of every token in the sentence. Specifically, we train a linear classifier to predict the correct tag from the contextualised representation of each token, while optionally also fine-tuning the encoder. Since most tokens do not belong to any entity span, the training data will inevitably be highly imbalanced. For this reason, during training, we ignore the majority of tokens that are outside of any entity span. Specifically, following \citet{hohenecker-etal-2020-systematic}, we only consider such tokens when they are immediately preceding or succeeding an entity span.

\section{Additional Analysis}
\label{app:modelOutputComparison}

\begin{table*}[t]
		\renewcommand{\arraystretch}{1.1}
		\footnotesize{
			\centering{\setlength\tabcolsep{2.4pt}
        \resizebox{\textwidth}{!}{
		\begin{tabular}{c@{\ \ \ }lccc}
		\toprule
         & \makecell[c]{\textbf{Sentence}} 
         & \textbf{Gold label}
         & \textbf{MLM-only}
         & \textbf{EnCore} \\ 
         
        \midrule
        (1) &
        \multirow{2}{*}{
        \makecell[l]{
        At the beginning of 1993 , six cities such as Zhuhai , Foshan , etc. also organized a \textbf{delegation} to \\ advertise in the US and Canada for students studying abroad.
        }
        } 
        & /organization 
        & 0.26 
        & 0.60 \\
        & 
        & /other 
        & 0.54 
        & 0.15 \\

        \midrule
        (2) &
        \multirow{2}{*}{
        \makecell[l]{
        Last year , its foreign exchange income was up to more than 2.1 billion US \textbf{dollars}, and in the first\\ half of this year exports again had new growth.
        }
        } 
        & /other 
        & 0.63
        & 0.97 \\
        & 
        & /other/currency 
        & 0.04
        & 0.98 \\

        \midrule
        (3) &
        \multirow{2}{*}{
        \makecell[l]{
        In 1997 , this plant made over 4,400 tons of Mao - tai ; with sales income exceeding 500 million\\ yuan RMB , and profit and taxes reaching 370 million \textbf{RMB} , both being the best levels in history.
        }
        } 
        & /other 
        & 0.31 
        & 0.94 \\
        & 
        & /other/currency 
        & 0.02 
        & 0.96 \\

        \midrule
        (4) &
        \multirow{2}{*}{
        \makecell[l]{
        In the near future , the Russian Tumen River Region Negotiation Conference will also be held in\\ \textbf{Vladivostok}.
        }
        } 
        & /location 
        & 0.25 
        & 0.98 \\
        & 
        & /location/city 
        & 0.07 
        & 0.73 \\
        \bottomrule
        \end{tabular}}}}
	\caption{Comparison of the confidence of the \emph{MLM-only} and \emph{EnCore} models (with \texttt{roberta-large}) on sample cases from the OntoNotes test set. The words in \textbf{bold} in the input sentences are the entity spans' head word. The \textbf{MLM-only} and \textbf{EnCore} columns indicate the confidence \steven{of} the \emph{MLM-only} and \emph{EnCore} models, respectively.}\label{tabComparisonModelOutputs}
\end{table*}

\paragraph{Prediction confidence}
In Table~\ref{tabComparisonModelOutputs}, we compare the confidence of the EnCore and MLM-only models for the gold label predictions. We observe that in the first example, EnCore more confidently predicts the label for \textit{delegation} as \textit{/organization} than MLM-only, which places \textit{delegation} in the more generic label class \textit{/other} with lower confidence.
In the second and third case, we observe that EnCore is more certain to label the currency terms \textit{dollars} and \textit{RMB} with the second-level label \textit{/other/currency} than with the more general first level label \textit{/other}, whereas MLM-only assigns a very low confidence to \textit{/other/currency}. A similar pattern can also be observed in the last example.
We have observed the same trend throughout the test set: EnCore consistently makes more confident predictions than MLM-only. This is especially evident for the second- and third-level labels.

\begin{figure*}[t]    
 \centering
 \includegraphics[width=\textwidth]{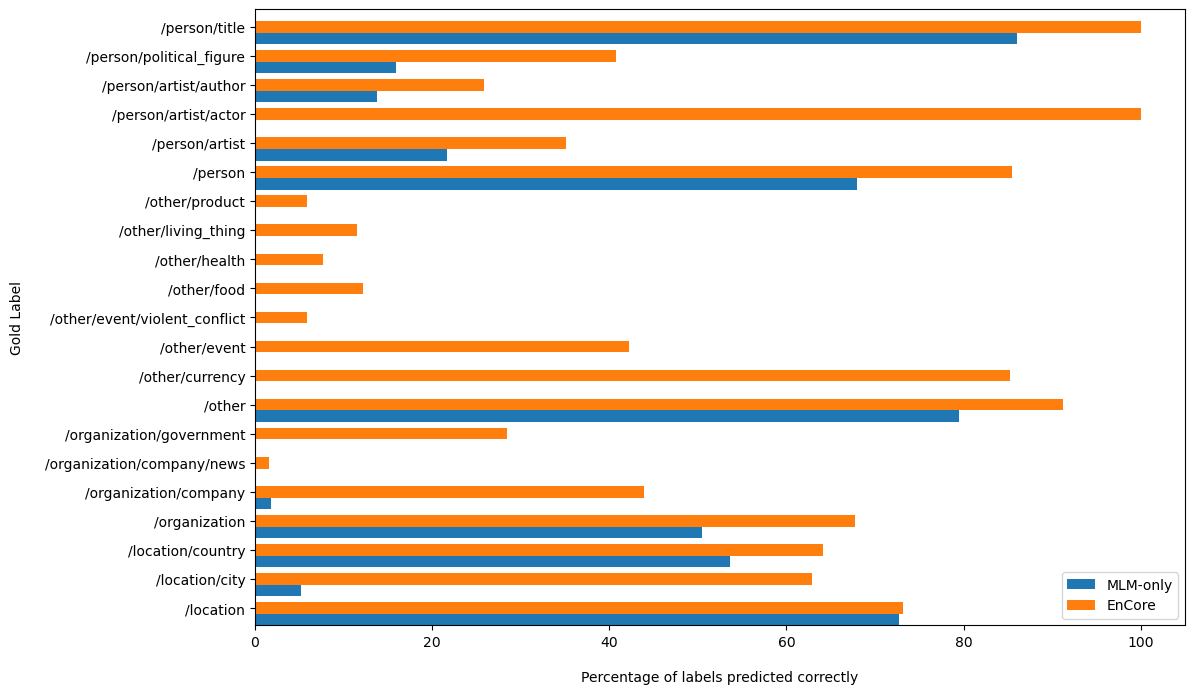}
    \caption{
    Comparison of the percentage of correct predictions per gold label by the MLM-only and EnCore models (with \texttt{roberta-large}) on the OntoNotes test set. The instances of a label that are accurately predicted are expressed as a percentage of the total number of occurrences of the corresponding gold label.
\label{fig:mlmEnCorePredictionsOntoNotes}}
\end{figure*}

\begin{figure*}[t]    
 \centering
 \includegraphics[width=\textwidth]{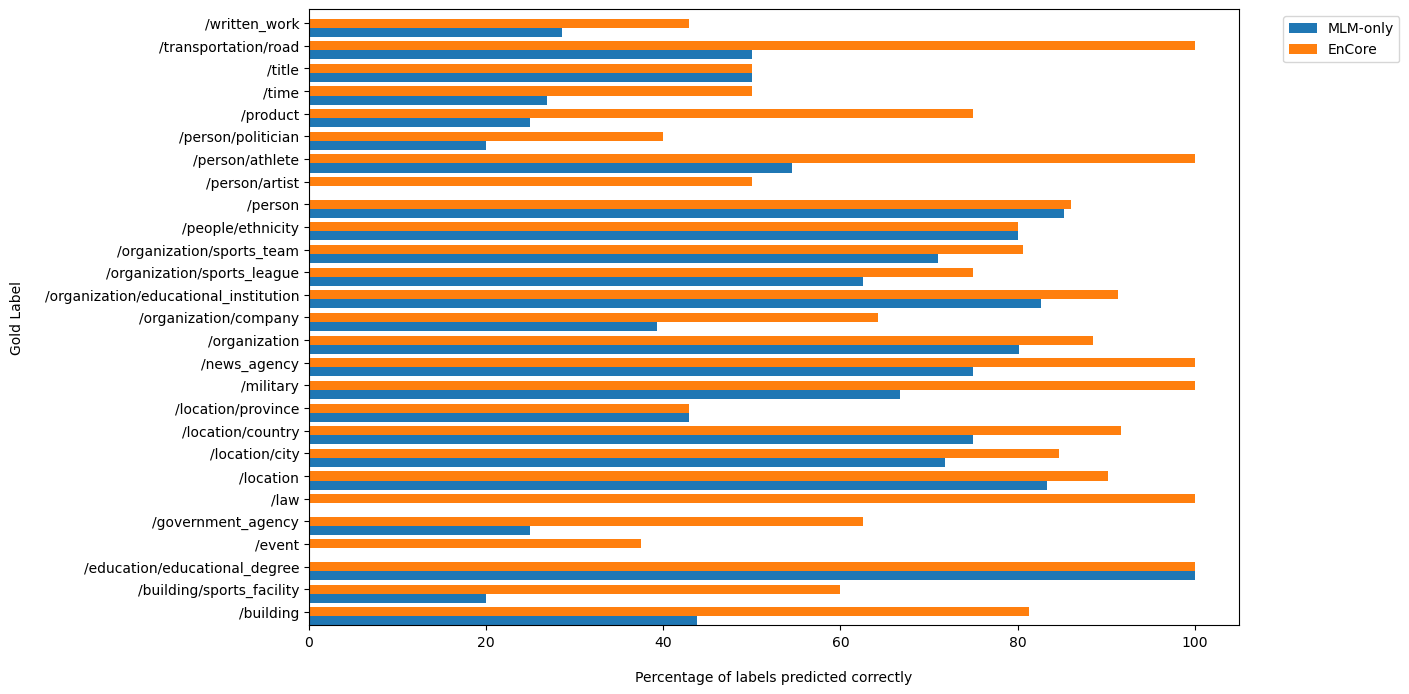}
    \caption{
    Comparison of the percentage of correct predictions per gold label by the MLM-only and EnCore models (with \texttt{roberta-large}) on the FIGER test set. The instances of a label that are accurately predicted are expressed as a percentage of the total number of occurrences of the corresponding gold label.
\label{fig:mlmEnCorePredictionsFiger}}
\end{figure*}

\paragraph{Breakdown by Label}
A closer examination of the model outputs in Figure~\ref{fig:mlmEnCorePredictionsOntoNotes} reveals that EnCore consistently beats the MLM-only model across all entity types. %As you go from first-level labels to second- and third-level labels, the distinction is still very noticeable. 
The OntoNotes test set, for example, contains 1130 \textit{/person} gold labels. MLM-only predicts only 67.96\% of these accurately, compared to 85.49\% for EnCore. As an example of a label at the second level, there are 74 \textit{/person/artist } gold labels in the test set; the MLM-only model correctly predicts 21.62\% of these, whereas EnCore correctly predicts 35.14\%. At the third level, there are 58 \textit{/person/artist/author} gold labels. The MLM-only model predicts only 13.79\% of them correctly, while EnCore predicts 25.86\% correctly. These patterns are consistently seen over the whole label set.
This is also true for the FIGER test set, as shown in Figure~\ref{fig:mlmEnCorePredictionsFiger}.

%% file: acl_latex.bbl
\begin{thebibliography}{50}
\expandafter\ifx\csname natexlab\endcsname\relax\def\natexlab#1{#1}\fi

\bibitem[{Baldini~Soares et~al.(2019)Baldini~Soares, FitzGerald, Ling, and Kwiatkowski}]{baldini-soares-etal-2019-matching}
Livio Baldini~Soares, Nicholas FitzGerald, Jeffrey Ling, and Tom Kwiatkowski. 2019.
\newblock \href {https://doi.org/10.18653/v1/P19-1279} {Matching the blanks: Distributional similarity for relation learning}.
\newblock In \emph{Proceedings of the 57th Annual Meeting of the Association for Computational Linguistics}, pages 2895--2905, Florence, Italy. Association for Computational Linguistics.

\bibitem[{Bekoulis et~al.(2018)Bekoulis, Deleu, Demeester, and Develder}]{bekoulis-etal-2018-adversarial}
Giannis Bekoulis, Johannes Deleu, Thomas Demeester, and Chris Develder. 2018.
\newblock \href {https://doi.org/10.18653/v1/D18-1307} {Adversarial training for multi-context joint entity and relation extraction}.
\newblock In \emph{Proceedings of the 2018 Conference on Empirical Methods in Natural Language Processing}, pages 2830--2836, Brussels, Belgium. Association for Computational Linguistics.

\bibitem[{Bollacker et~al.(2008)Bollacker, Evans, Paritosh, Sturge, and Taylor}]{bollacker2008freebase}
Kurt Bollacker, Colin Evans, Praveen Paritosh, Tim Sturge, and Jamie Taylor. 2008.
\newblock Freebase: a collaboratively created graph database for structuring human knowledge.
\newblock In \emph{Proceedings of the 2008 ACM SIGMOD international conference on Management of data}, pages 1247--1250.

\bibitem[{Choi et~al.(2018)Choi, Levy, Choi, and Zettlemoyer}]{choi-etal-2018-ultra}
Eunsol Choi, Omer Levy, Yejin Choi, and Luke Zettlemoyer. 2018.
\newblock \href {https://doi.org/10.18653/v1/P18-1009} {Ultra-fine entity typing}.
\newblock In \emph{Proceedings of the 56th Annual Meeting of the Association for Computational Linguistics (Volume 1: Long Papers)}, pages 87--96, Melbourne, Australia. Association for Computational Linguistics.

\bibitem[{Dai et~al.(2021)Dai, Song, and Wang}]{dai-etal-2021-ultra}
Hongliang Dai, Yangqiu Song, and Haixun Wang. 2021.
\newblock \href {https://doi.org/10.18653/v1/2021.acl-long.141} {Ultra-fine entity typing with weak supervision from a masked language model}.
\newblock In \emph{Proceedings of the 59th Annual Meeting of the Association for Computational Linguistics and the 11th International Joint Conference on Natural Language Processing (Volume 1: Long Papers)}, pages 1790--1799, Online. Association for Computational Linguistics.

\bibitem[{Devlin et~al.(2019)Devlin, Chang, Lee, and Toutanova}]{devlin-etal-2019-bert}
Jacob Devlin, Ming-Wei Chang, Kenton Lee, and Kristina Toutanova. 2019.
\newblock \href {https://doi.org/10.18653/v1/N19-1423} {{BERT}: Pre-training of deep bidirectional transformers for language understanding}.
\newblock In \emph{Proceedings of the 2019 Conference of the North {A}merican Chapter of the Association for Computational Linguistics: Human Language Technologies, Volume 1 (Long and Short Papers)}, pages 4171--4186, Minneapolis, Minnesota. Association for Computational Linguistics.

\bibitem[{Gao et~al.(2021)Gao, Yao, and Chen}]{gao-etal-2021-simcse}
Tianyu Gao, Xingcheng Yao, and Danqi Chen. 2021.
\newblock \href {https://doi.org/10.18653/v1/2021.emnlp-main.552} {{S}im{CSE}: Simple contrastive learning of sentence embeddings}.
\newblock In \emph{Proceedings of the 2021 Conference on Empirical Methods in Natural Language Processing}, pages 6894--6910, Online and Punta Cana, Dominican Republic. Association for Computational Linguistics.

\bibitem[{Gillick et~al.(2014)Gillick, Lazic, Ganchev, Kirchner, and Huynh}]{DBLP:journals/corr/GillickLGKH14}
Dan Gillick, Nevena Lazic, Kuzman Ganchev, Jesse Kirchner, and David Huynh. 2014.
\newblock \href {http://arxiv.org/abs/1412.1820} {Context-dependent fine-grained entity type tagging}.
\newblock \emph{CoRR}, abs/1412.1820.

\bibitem[{Han et~al.(2021)Han, Cheng, and Lu}]{han-etal-2021-exploring}
Jiale Han, Bo~Cheng, and Wei Lu. 2021.
\newblock \href {https://doi.org/10.18653/v1/2021.emnlp-main.204} {Exploring task difficulty for few-shot relation extraction}.
\newblock In \emph{Proceedings of the 2021 Conference on Empirical Methods in Natural Language Processing}, pages 2605--2616, Online and Punta Cana, Dominican Republic. Association for Computational Linguistics.

\bibitem[{Han et~al.(2023)Han, Peng, Yang, Wang, Liu, and Wan}]{DBLP:journals/corr/abs-2305-14450}
Ridong Han, Tao Peng, Chaohao Yang, Benyou Wang, Lu~Liu, and Xiang Wan. 2023.
\newblock \href {https://doi.org/10.48550/arXiv.2305.14450} {Is information extraction solved by chatgpt? an analysis of performance, evaluation criteria, robustness and errors}.
\newblock \emph{CoRR}, abs/2305.14450.

\bibitem[{Hohenecker et~al.(2020)Hohenecker, Mtumbuka, Kocijan, and Lukasiewicz}]{hohenecker-etal-2020-systematic}
Patrick Hohenecker, Frank Mtumbuka, Vid Kocijan, and Thomas Lukasiewicz. 2020.
\newblock \href {https://doi.org/10.18653/v1/2020.emnlp-main.690} {Systematic comparison of neural architectures and training approaches for open information extraction}.
\newblock In \emph{Proceedings of the 2020 Conference on Empirical Methods in Natural Language Processing (EMNLP)}, pages 8554--8565, Online. Association for Computational Linguistics.

\bibitem[{Hu et~al.(2021)Hu, Qiao, Xing, and Peng}]{DBLP:journals/access/HuQXP21}
Yanfeng Hu, Xue Qiao, Luo Xing, and Chen Peng. 2021.
\newblock \href {https://doi.org/10.1109/ACCESS.2020.3046787} {Diversified semantic attention model for fine-grained entity typing}.
\newblock \emph{{IEEE} Access}, 9:2251--2265.

\bibitem[{Huang et~al.(2022)Huang, Li, Xu, and Chen}]{huang-etal-2022-unified}
James~Y. Huang, Bangzheng Li, Jiashu Xu, and Muhao Chen. 2022.
\newblock \href {https://doi.org/10.18653/v1/2022.naacl-main.190} {Unified semantic typing with meaningful label inference}.
\newblock In \emph{Proceedings of the 2022 Conference of the North American Chapter of the Association for Computational Linguistics: Human Language Technologies}, pages 2642--2654, Seattle, United States. Association for Computational Linguistics.

\bibitem[{Joshi et~al.(2020)Joshi, Chen, Liu, Weld, Zettlemoyer, and Levy}]{joshi-etal-2020-spanbert}
Mandar Joshi, Danqi Chen, Yinhan Liu, Daniel~S. Weld, Luke Zettlemoyer, and Omer Levy. 2020.
\newblock \href {https://doi.org/10.1162/tacl_a_00300} {{S}pan{BERT}: Improving pre-training by representing and predicting spans}.
\newblock \emph{Transactions of the Association for Computational Linguistics}, 8:64--77.

\bibitem[{Li et~al.(2022{\natexlab{a}})Li, Yin, and Chen}]{li-etal-2022-ultra}
Bangzheng Li, Wenpeng Yin, and Muhao Chen. 2022{\natexlab{a}}.
\newblock \href {https://doi.org/10.1162/tacl_a_00479} {Ultra-fine entity typing with indirect supervision from natural language inference}.
\newblock \emph{Transactions of the Association for Computational Linguistics}, 10:607--622.

\bibitem[{Li et~al.(2022{\natexlab{b}})Li, Shang, and McAuley}]{li-etal-2022-uctopic}
Jiacheng Li, Jingbo Shang, and Julian McAuley. 2022{\natexlab{b}}.
\newblock \href {https://doi.org/10.18653/v1/2022.acl-long.426} {{UCT}opic: Unsupervised contrastive learning for phrase representations and topic mining}.
\newblock In \emph{Proceedings of the 60th Annual Meeting of the Association for Computational Linguistics (Volume 1: Long Papers)}, pages 6159--6169, Dublin, Ireland. Association for Computational Linguistics.

\bibitem[{Li et~al.(2021)Li, Chen, Wang, and Li}]{DBLP:conf/ijcai/Li0WL21}
Jinqing Li, Xiaojun Chen, Dakui Wang, and Yuwei Li. 2021.
\newblock \href {https://doi.org/10.24963/ijcai.2021/529} {Enhancing label representations with relational inductive bias constraint for fine-grained entity typing}.
\newblock In \emph{Proceedings of the Thirtieth International Joint Conference on Artificial Intelligence, {IJCAI} 2021, Virtual Event / Montreal, Canada, 19-27 August 2021}, pages 3843--3849. ijcai.org.

\bibitem[{Li et~al.(2023)Li, Bouraoui, and Schockaert}]{DBLP:journals/corr/abs-2305-12802}
Na~Li, Zied Bouraoui, and Steven Schockaert. 2023.
\newblock \href {https://doi.org/10.48550/arXiv.2305.12802} {Ultra-fine entity typing with prior knowledge about labels: {A} simple clustering based strategy}.
\newblock \emph{CoRR}, abs/2305.12802.

\bibitem[{Li and Ji(2014)}]{li-ji-2014-incremental}
Qi~Li and Heng Ji. 2014.
\newblock \href {https://doi.org/10.3115/v1/P14-1038} {Incremental joint extraction of entity mentions and relations}.
\newblock In \emph{Proceedings of the 52nd Annual Meeting of the Association for Computational Linguistics (Volume 1: Long Papers)}, pages 402--412, Baltimore, Maryland. Association for Computational Linguistics.

\bibitem[{Ling and Weld(2012)}]{DBLP:conf/aaai/LingW12}
Xiao Ling and Daniel~S. Weld. 2012.
\newblock \href {http://www.aaai.org/ocs/index.php/AAAI/AAAI12/paper/view/5152} {Fine-grained entity recognition}.
\newblock In \emph{Proceedings of the Twenty-Sixth {AAAI} Conference on Artificial Intelligence, July 22-26, 2012, Toronto, Ontario, Canada}. {AAAI} Press.

\bibitem[{Liu et~al.(2021{\natexlab{a}})Liu, Vuli{\'c}, Korhonen, and Collier}]{liu-etal-2021-fast}
Fangyu Liu, Ivan Vuli{\'c}, Anna Korhonen, and Nigel Collier. 2021{\natexlab{a}}.
\newblock \href {https://doi.org/10.18653/v1/2021.emnlp-main.109} {Fast, effective, and self-supervised: Transforming masked language models into universal lexical and sentence encoders}.
\newblock In \emph{Proceedings of the 2021 Conference on Empirical Methods in Natural Language Processing}, pages 1442--1459, Online and Punta Cana, Dominican Republic. Association for Computational Linguistics.

\bibitem[{Liu et~al.(2021{\natexlab{b}})Liu, Liu, Collier, Korhonen, and Vuli{\'c}}]{liu-etal-2021-mirrorwic}
Qianchu Liu, Fangyu Liu, Nigel Collier, Anna Korhonen, and Ivan Vuli{\'c}. 2021{\natexlab{b}}.
\newblock \href {https://doi.org/10.18653/v1/2021.conll-1.44} {{M}irror{W}i{C}: On eliciting word-in-context representations from pretrained language models}.
\newblock In \emph{Proceedings of the 25th Conference on Computational Natural Language Learning}, pages 562--574, Online. Association for Computational Linguistics.

\bibitem[{Loshchilov and Hutter(2019)}]{loshchilov2018decoupled}
Ilya Loshchilov and Frank Hutter. 2019.
\newblock Decoupled weight decay regularization.
\newblock In \emph{International Conference on Learning Representations}.

\bibitem[{Luan et~al.(2018)Luan, He, Ostendorf, and Hajishirzi}]{luan-etal-2018-multi}
Yi~Luan, Luheng He, Mari Ostendorf, and Hannaneh Hajishirzi. 2018.
\newblock \href {https://doi.org/10.18653/v1/D18-1360} {Multi-task identification of entities, relations, and coreference for scientific knowledge graph construction}.
\newblock In \emph{Proceedings of the 2018 Conference on Empirical Methods in Natural Language Processing}, pages 3219--3232, Brussels, Belgium. Association for Computational Linguistics.

\bibitem[{Onoe et~al.(2021)Onoe, Boratko, McCallum, and Durrett}]{onoe-etal-2021-modeling}
Yasumasa Onoe, Michael Boratko, Andrew McCallum, and Greg Durrett. 2021.
\newblock \href {https://doi.org/10.18653/v1/2021.acl-long.160} {Modeling fine-grained entity types with box embeddings}.
\newblock In \emph{Proceedings of the 59th Annual Meeting of the Association for Computational Linguistics and the 11th International Joint Conference on Natural Language Processing (Volume 1: Long Papers)}, pages 2051--2064, Online. Association for Computational Linguistics.

\bibitem[{Onoe and Durrett(2019)}]{onoe-durrett-2019-learning}
Yasumasa Onoe and Greg Durrett. 2019.
\newblock \href {https://doi.org/10.18653/v1/N19-1250} {Learning to denoise distantly-labeled data for entity typing}.
\newblock In \emph{Proceedings of the 2019 Conference of the North {A}merican Chapter of the Association for Computational Linguistics: Human Language Technologies, Volume 1 (Long and Short Papers)}, pages 2407--2417, Minneapolis, Minnesota. Association for Computational Linguistics.

\bibitem[{Onoe and Durrett(2020)}]{onoe-durrett-2020-interpretable}
Yasumasa Onoe and Greg Durrett. 2020.
\newblock \href {https://doi.org/10.18653/v1/2020.findings-emnlp.54} {Interpretable entity representations through large-scale typing}.
\newblock In \emph{Findings of the Association for Computational Linguistics: EMNLP 2020}, pages 612--624, Online. Association for Computational Linguistics.

\bibitem[{Pan et~al.(2022)Pan, Wei, and Zhu}]{DBLP:conf/ijcai/Pan0022}
Weiran Pan, Wei Wei, and Feida Zhu. 2022.
\newblock \href {https://doi.org/10.24963/ijcai.2022/599} {Automatic noisy label correction for fine-grained entity typing}.
\newblock In \emph{Proceedings of the Thirty-First International Joint Conference on Artificial Intelligence, {IJCAI} 2022, Vienna, Austria, 23-29 July 2022}, pages 4317--4323. ijcai.org.

\bibitem[{Peng et~al.(2020)Peng, Gao, Han, Lin, Li, Liu, Sun, and Zhou}]{peng-etal-2020-learning}
Hao Peng, Tianyu Gao, Xu~Han, Yankai Lin, Peng Li, Zhiyuan Liu, Maosong Sun, and Jie Zhou. 2020.
\newblock \href {https://doi.org/10.18653/v1/2020.emnlp-main.298} {{L}earning from {C}ontext or {N}ames? {A}n {E}mpirical {S}tudy on {N}eural {R}elation {E}xtraction}.
\newblock In \emph{Proceedings of the 2020 Conference on Empirical Methods in Natural Language Processing (EMNLP)}, pages 3661--3672, Online. Association for Computational Linguistics.

\bibitem[{Peters et~al.(2018)Peters, Neumann, Iyyer, Gardner, Clark, Lee, and Zettlemoyer}]{peters-etal-2018-deep}
Matthew~E. Peters, Mark Neumann, Mohit Iyyer, Matt Gardner, Christopher Clark, Kenton Lee, and Luke Zettlemoyer. 2018.
\newblock \href {https://doi.org/10.18653/v1/N18-1202} {Deep contextualized word representations}.
\newblock In \emph{Proceedings of the 2018 Conference of the North {A}merican Chapter of the Association for Computational Linguistics: Human Language Technologies, Volume 1 (Long Papers)}, pages 2227--2237, New Orleans, Louisiana. Association for Computational Linguistics.

\bibitem[{Peters et~al.(2019)Peters, Neumann, Logan, Schwartz, Joshi, Singh, and Smith}]{peters-etal-2019-knowledge}
Matthew~E. Peters, Mark Neumann, Robert Logan, Roy Schwartz, Vidur Joshi, Sameer Singh, and Noah~A. Smith. 2019.
\newblock \href {https://doi.org/10.18653/v1/D19-1005} {Knowledge enhanced contextual word representations}.
\newblock In \emph{Proceedings of the 2019 Conference on Empirical Methods in Natural Language Processing and the 9th International Joint Conference on Natural Language Processing (EMNLP-IJCNLP)}, pages 43--54, Hong Kong, China. Association for Computational Linguistics.

\bibitem[{Ren et~al.(2016)Ren, He, Qu, Voss, Ji, and Han}]{DBLP:conf/kdd/RenHQVJH16}
Xiang Ren, Wenqi He, Meng Qu, Clare~R. Voss, Heng Ji, and Jiawei Han. 2016.
\newblock \href {https://doi.org/10.1145/2939672.2939822} {Label noise reduction in entity typing by heterogeneous partial-label embedding}.
\newblock In \emph{Proceedings of the 22nd {ACM} {SIGKDD} International Conference on Knowledge Discovery and Data Mining, San Francisco, CA, USA, August 13-17, 2016}, pages 1825--1834. {ACM}.

\bibitem[{van~den Oord et~al.(2018)van~den Oord, Li, and Vinyals}]{DBLP:journals/corr/abs-1807-03748}
A{\"{a}}ron van~den Oord, Yazhe Li, and Oriol Vinyals. 2018.
\newblock \href {http://arxiv.org/abs/1807.03748} {Representation learning with contrastive predictive coding}.
\newblock \emph{CoRR}, abs/1807.03748.

\bibitem[{Varkel and Globerson(2020)}]{varkel-globerson-2020-pre}
Yuval Varkel and Amir Globerson. 2020.
\newblock \href {https://doi.org/10.18653/v1/2020.emnlp-main.687} {Pre-training mention representations in coreference models}.
\newblock In \emph{Proceedings of the 2020 Conference on Empirical Methods in Natural Language Processing (EMNLP)}, pages 8534--8540, Online. Association for Computational Linguistics.

\bibitem[{Wadden et~al.(2019)Wadden, Wennberg, Luan, and Hajishirzi}]{wadden-etal-2019-entity}
David Wadden, Ulme Wennberg, Yi~Luan, and Hannaneh Hajishirzi. 2019.
\newblock \href {https://doi.org/10.18653/v1/D19-1585} {Entity, relation, and event extraction with contextualized span representations}.
\newblock In \emph{Proceedings of the 2019 Conference on Empirical Methods in Natural Language Processing and the 9th International Joint Conference on Natural Language Processing (EMNLP-IJCNLP)}, pages 5784--5789, Hong Kong, China. Association for Computational Linguistics.

\bibitem[{Wan et~al.(2022)Wan, Cheng, Liu, Mao, Song, and Kurohashi}]{DBLP:journals/corr/abs-2205-08770}
Zhen Wan, Fei Cheng, Qianying Liu, Zhuoyuan Mao, Haiyue Song, and Sadao Kurohashi. 2022.
\newblock \href {https://doi.org/10.48550/arXiv.2205.08770} {Relation extraction with weighted contrastive pre-training on distant supervision}.
\newblock \emph{CoRR}, abs/2205.08770.

\bibitem[{Wang and Lu(2020)}]{wang-lu-2020-two}
Jue Wang and Wei Lu. 2020.
\newblock \href {https://doi.org/10.18653/v1/2020.emnlp-main.133} {Two are better than one: Joint entity and relation extraction with table-sequence encoders}.
\newblock In \emph{Proceedings of the 2020 Conference on Empirical Methods in Natural Language Processing (EMNLP)}, pages 1706--1721, Online. Association for Computational Linguistics.

\bibitem[{Wang et~al.(2021{\natexlab{a}})Wang, Tang, Duan, Wei, Huang, Ji, Cao, Jiang, and Zhou}]{wang-etal-2021-k}
Ruize Wang, Duyu Tang, Nan Duan, Zhongyu Wei, Xuanjing Huang, Jianshu Ji, Guihong Cao, Daxin Jiang, and Ming Zhou. 2021{\natexlab{a}}.
\newblock \href {https://doi.org/10.18653/v1/2021.findings-acl.121} {{K-Adapter}: {I}nfusing {K}nowledge into {P}re-{T}rained {M}odels with {A}dapters}.
\newblock In \emph{Findings of the Association for Computational Linguistics: ACL-IJCNLP 2021}, pages 1405--1418, Online. Association for Computational Linguistics.

\bibitem[{Wang et~al.(2021{\natexlab{b}})Wang, Thompson, and Iyyer}]{wang-etal-2021-phrase}
Shufan Wang, Laure Thompson, and Mohit Iyyer. 2021{\natexlab{b}}.
\newblock \href {https://doi.org/10.18653/v1/2021.emnlp-main.846} {Phrase-{BERT}: Improved phrase embeddings from {BERT} with an application to corpus exploration}.
\newblock In \emph{Proceedings of the 2021 Conference on Empirical Methods in Natural Language Processing}, pages 10837--10851, Online and Punta Cana, Dominican Republic. Association for Computational Linguistics.

\bibitem[{Wang et~al.(2022)Wang, Zhang, Xu, Wu, and Xiao}]{wang-etal-2022-rcl}
Shusen Wang, Bosen Zhang, Yajing Xu, Yanan Wu, and Bo~Xiao. 2022.
\newblock \href {https://doi.org/10.18653/v1/2022.findings-naacl.188} {{RCL}: Relation contrastive learning for zero-shot relation extraction}.
\newblock In \emph{Findings of the Association for Computational Linguistics: NAACL 2022}, pages 2456--2468, Seattle, United States. Association for Computational Linguistics.

\bibitem[{Wang et~al.(2023)Wang, Zhou, Zu, Xia, Chen, Zhang, Zheng, Ye, Zhang, Gui, Kang, Yang, Li, and Du}]{DBLP:journals/corr/abs-2304-08085}
Xiao Wang, Weikang Zhou, Can Zu, Han Xia, Tianze Chen, Yuansen Zhang, Rui Zheng, Junjie Ye, Qi~Zhang, Tao Gui, Jihua Kang, Jingsheng Yang, Siyuan Li, and Chunsai Du. 2023.
\newblock \href {https://doi.org/10.48550/arXiv.2304.08085} {{InstructUIE}: Multi-task instruction tuning for unified information extraction}.
\newblock \emph{CoRR}, abs/2304.08085.

\bibitem[{Wang et~al.(2021{\natexlab{c}})Wang, Gao, Zhu, Zhang, Liu, Li, and Tang}]{wang-etal-2021-kepler}
Xiaozhi Wang, Tianyu Gao, Zhaocheng Zhu, Zhengyan Zhang, Zhiyuan Liu, Juanzi Li, and Jian Tang. 2021{\natexlab{c}}.
\newblock \href {https://doi.org/10.1162/tacl_a_00360} {{KEPLER}: A unified model for knowledge embedding and pre-trained language representation}.
\newblock \emph{Transactions of the Association for Computational Linguistics}, 9:176--194.

\bibitem[{Wang et~al.(2020)Wang, Sun, Wu, Yan, Gao, and Xie}]{wang-etal-2020-pre}
Yijun Wang, Changzhi Sun, Yuanbin Wu, Junchi Yan, Peng Gao, and Guotong Xie. 2020.
\newblock \href {https://doi.org/10.18653/v1/2020.emnlp-main.132} {Pre-training entity relation encoder with intra-span and inter-span information}.
\newblock In \emph{Proceedings of the 2020 Conference on Empirical Methods in Natural Language Processing (EMNLP)}, pages 1692--1705, Online. Association for Computational Linguistics.

\bibitem[{Wang et~al.(2021{\natexlab{d}})Wang, Sun, Wu, Zhou, Li, and Yan}]{wang-etal-2021-unire}
Yijun Wang, Changzhi Sun, Yuanbin Wu, Hao Zhou, Lei Li, and Junchi Yan. 2021{\natexlab{d}}.
\newblock \href {https://doi.org/10.18653/v1/2021.acl-long.19} {{U}ni{RE}: A unified label space for entity relation extraction}.
\newblock In \emph{Proceedings of the 59th Annual Meeting of the Association for Computational Linguistics and the 11th International Joint Conference on Natural Language Processing (Volume 1: Long Papers)}, pages 220--231, Online. Association for Computational Linguistics.

\bibitem[{Xu et~al.(2021)Xu, Guo, Tang, Su, Shou, Gong, Zhong, Quan, Jiang, and Duan}]{xu-etal-2021-syntax}
Zenan Xu, Daya Guo, Duyu Tang, Qinliang Su, Linjun Shou, Ming Gong, Wanjun Zhong, Xiaojun Quan, Daxin Jiang, and Nan Duan. 2021.
\newblock \href {https://doi.org/10.18653/v1/2021.acl-long.420} {Syntax-enhanced pre-trained model}.
\newblock In \emph{Proceedings of the 59th Annual Meeting of the Association for Computational Linguistics and the 11th International Joint Conference on Natural Language Processing (Volume 1: Long Papers)}, pages 5412--5422, Online. Association for Computational Linguistics.

\bibitem[{Yamada et~al.(2020)Yamada, Asai, Shindo, Takeda, and Matsumoto}]{yamada-etal-2020-luke}
Ikuya Yamada, Akari Asai, Hiroyuki Shindo, Hideaki Takeda, and Yuji Matsumoto. 2020.
\newblock \href {https://doi.org/10.18653/v1/2020.emnlp-main.523} {{LUKE}: Deep contextualized entity representations with entity-aware self-attention}.
\newblock In \emph{Proceedings of the 2020 Conference on Empirical Methods in Natural Language Processing (EMNLP)}, pages 6442--6454, Online. Association for Computational Linguistics.

\bibitem[{Ye et~al.(2022)Ye, Lin, Li, and Sun}]{ye-etal-2022-packed}
Deming Ye, Yankai Lin, Peng Li, and Maosong Sun. 2022.
\newblock \href {https://doi.org/10.18653/v1/2022.acl-long.337} {Packed levitated marker for entity and relation extraction}.
\newblock In \emph{Proceedings of the 60th Annual Meeting of the Association for Computational Linguistics (Volume 1: Long Papers)}, pages 4904--4917, Dublin, Ireland. Association for Computational Linguistics.

\bibitem[{Zhang et~al.(2019)Zhang, Han, Liu, Jiang, Sun, and Liu}]{zhang-etal-2019-ernie}
Zhengyan Zhang, Xu~Han, Zhiyuan Liu, Xin Jiang, Maosong Sun, and Qun Liu. 2019.
\newblock \href {https://doi.org/10.18653/v1/P19-1139} {{ERNIE}: Enhanced language representation with informative entities}.
\newblock In \emph{Proceedings of the 57th Annual Meeting of the Association for Computational Linguistics}, pages 1441--1451, Florence, Italy. Association for Computational Linguistics.

\bibitem[{Zhong and Chen(2021)}]{zhong-chen-2021-frustratingly}
Zexuan Zhong and Danqi Chen. 2021.
\newblock \href {https://doi.org/10.18653/v1/2021.naacl-main.5} {A frustratingly easy approach for entity and relation extraction}.
\newblock In \emph{Proceedings of the 2021 Conference of the North American Chapter of the Association for Computational Linguistics: Human Language Technologies}, pages 50--61, Online. Association for Computational Linguistics.

\bibitem[{Zuo et~al.(2022)Zuo, Liang, Jing, Zeng, Fang, and Luo}]{zuo-etal-2022-type}
Xinyu Zuo, Haijin Liang, Ning Jing, Shuang Zeng, Zhou Fang, and Yu~Luo. 2022.
\newblock \href {https://aclanthology.org/2022.coling-1.212} {Type-enriched hierarchical contrastive strategy for fine-grained entity typing}.
\newblock In \emph{Proceedings of the 29th International Conference on Computational Linguistics}, pages 2405--2417, Gyeongju, Republic of Korea. International Committee on Computational Linguistics.

\end{thebibliography}
